\documentclass[conference]{IEEEtran}
\IEEEoverridecommandlockouts

\usepackage{amsmath,amsfonts,amsthm}
\usepackage{algorithmic}
\usepackage{algorithm}
\usepackage{array}
\usepackage[caption=false,font=normalsize,labelfont=sf,textfont=sf]{subfig}
\usepackage{textcomp}
\usepackage{stfloats}
\usepackage{url}
\usepackage{verbatim}
\usepackage{graphicx}
\usepackage{cite}
\usepackage{enumitem}
\usepackage{style/macros}
\usepackage{bm}
\usepackage{booktabs}
\usepackage{multirow}
\usepackage{xcolor}
\usepackage{tikz}

\newtheorem{definition}{Definition}

\newtheorem{theorem}{Theorem}
\newtheorem{assumption}{Assumption}

\newcommand{\avg}[1]{ \lim_{T \to \infty} \frac{1}{T} \int_{-T/2}^{T/2} #1 dt}
\def\inW{A}
\def\outW{B}
\newcommand{\Linf}{\calL_\infty}
\newcommand{\Lfin}{\calL_\sqcap}

\begin{document}
\bstctlcite{IEEEexample:BSTcontrol}

\title{Solving Large-scale Spatial Problems with Convolutional Neural Networks\\
}

\author{\IEEEauthorblockN{1\textsuperscript{st} Damian Owerko}
    \IEEEauthorblockA{\textit{Electrical and Systems Engineering} \\
        \textit{University of Pennsylvania}\\
        Philadelphia, USA \\
        owerko@seas.upenn.edu}
    \and
    \IEEEauthorblockN{2\textsuperscript{nd} Charilaos I. Kanatsoulis}
    \IEEEauthorblockA{\textit{Electrical and Systems Engineering} \\
        \textit{University of Pennsylvania}\\
        Philadelphia, USA \\
        kanac@seas.upenn.edu}
    \and
    \IEEEauthorblockN{3\textsuperscript{rd} Alejandro Ribeiro}
    \IEEEauthorblockA{\textit{Electrical and Systems Engineering} \\
        \textit{University of Pennsylvania}\\
        Philadelphia, USA \\
        aribeiro@seas.upenn.edu}
}

\maketitle

\begin{abstract}
    Over the past decade, deep learning research has been accelerated by increasingly powerful hardware, which facilitated rapid growth in the model complexity and the amount of data ingested. This is becoming unsustainable and therefore refocusing on efficiency is necessary. In this paper, we employ transfer learning to improve training efficiency for large-scale spatial problems. We propose that a convolutional neural network (CNN) can be trained on small windows of signals, but evaluated on arbitrarily large signals with little to no performance degradation, and provide a theoretical bound on the resulting generalization error. Our proof leverages shift-equivariance of CNNs, a property that is underexploited in transfer learning. The theoretical results are experimentally supported in the context of mobile infrastructure on demand (MID). The proposed approach is able to tackle MID at large scales with hundreds of agents, which was computationally intractable prior to this work.

\end{abstract}


\section{Introduction}

Over the past decade, there has been a rapid advancement in machine learning (ML), particularly in deep learning, which has produced state-of-the-art results in a wide range of applications \cite{Shinde18-ReviewMachine, Gu18-RecentAdvances, Lin21-SurveyTransformers}. This progress has been fueled by increasingly powerful hardware \cite{Shinde18-ReviewMachine, Gu18-RecentAdvances} that has enabled the processing of larger datasets \cite{Najafabadi15-DeepLearning} and the training of deep learning models with more parameters. Theoretical evidence \cite{Du18-Power, Soltanolkotabi19-Theoretical} and empirical evidence \cite{Kaplan20-ScalingLawsNeural, Hoffmann22-EmpiricalAnalysis} suggest that using overparametrized models and larger datasets benefits neural network training. Large language models, such as GPT-3, with 175 billion parameters trained on a dataset of approximately 374 billion words, represent a new extreme in this trend \cite{Brown20-LanguageModelsAre, Rae22-ScalingLanguageModels, Smith22-UsingDeepSpeed, Touvron23-LLaMAOpen}. However, the trend of increasing model complexity and dataset size is not sustainable in the long term due to diminishing returns on costs of computation and data acquisition \cite{Thompson22-Computational, Patterson21-CarbonEmissions}. Moreover, some applications lack data availability, making this strategy impossible. Therefore, it is necessary to refocus on efficiency and explore more sustainable ML approaches.

Transfer learning \cite{Zhuang21-ComprehensiveSurvey, Tan18-SurveyDeepTransfer, Ribani19-SurveyTransfer, Zhu22-TransferLearningDeep} is a powerful tool for efficient and sustainable ML. It refers to a set of methodologies to apply knowledge learned from a source domain to a different target domain. For example, in \cite{Shin16-DeepConvolutional} the authors demonstrate that it is consistently beneficial to pre-train a convolutional neural network (CNN) on ImageNet before fine-tuning on medical images. In this case, transfer learning is especially beneficial because of the unavailability of large medical image datasets.

CNNs are one of the most popular deep learning architectures \cite{Gu18-RecentAdvances}, especially for image classification \cite{Rawat17-DeepConvolutional}. Although initially used for image processing, they have proven useful for a wide variety of other signals such as text, audio, weather, ECG data, traffic data and many others \cite{Gu18-RecentAdvances, Li22-SurveyConvolutional, Alipour19-RobustPixelLevel}. Shift-equivariance is an interesting property of CNNs. When there are no dilations, any translation of the input to the CNN will also translate the output by the same amount. Previous works focus on leveraging this property to achieve translation invariant image classification \cite{Zhang19-MakingConvolutional, Chaman21-TrulyShiftinvariant}. However, it is difficult to exploit shift-equivariance
for small images with deep architectures \cite{Azulay18-WhyDeep, SemihKayhan20-Translation}. Nevertheless, our work shows that shift-equivariance is fundamental for efficient large-scale image-to-image regression tasks, as we explain below.


In this paper, we use CNNs and transfer learning to tackle large-scale spatial problems. In particular, we leverage the shift-equivariance property of CNNs to efficiently train when the input-output signals are jointly stationary. Our analysis uses stochastic process theory to provide a bound on the generalization error of CNNs. The derived bound implies that a CNN can be trained on small signal windows, yet evaluated on arbitrarily large windows with minimal performance loss. Following, our theoretical result, we propose to recast spatial problems as image-to-image prediction tasks and use CNNs to solve them on a large scale. The proposed framework is applied to mobile infrastructure on demand (MID) tasks \cite{Mox20-MobileWirelessNetwork}. Our experimental results showcase that transfer learning with CNNs can tackle MID at scales that were previously considered intractable. Our main contributions are summarized as follows.
\begin{enumerate}[label=\textbf{(C\arabic*)}]
    \item Provide a bound on CNN generalization error after training on a small window and executing on arbitrarily large signals.
    \item Propose how to reinterpret large-scale spatial problems as image-to-image tasks.
    \item Demonstrate the proposed method by solving the MID problem at scale.
\end{enumerate}

\noindent\textbf{Notation:} We denote a stochastic process as \( \{ X(t) \}_t \) where each \( X(t) \) is a random variable. Correspondingly, \(X\) denotes a random signal -- a random element in a function space. Sets \(\calX\) are denoted by calligraphic letters. Lowercase boldface symbols \(\bfx\) are vectors while uppercase ones \(\bfX\) are matrices with \(\bfX_{ij}\) being their elements. \(\bm\Phi\) is an exception that symbolizes an ML model by convention.

\section{Learning to process stationary signals}


Commonly in ML the task is to find a model \(\bm\Phi(\cdot)\) which minimizes the expected mean squared error (MSE) between an input \( X \) and a desired output \( Y \). Let \(\bm\Phi(X)(t)\) be the output of the model at time \(t\) given the signal \(X\).
\begin{equation}\label{eq:mse_general}
    \min_{\bm\Phi} \E{ \avg{|\bm\Phi(X)(t) - Y(t)|^2}}
\end{equation}
\eqref{eq:mse_general} expresses this mean squared error minimization problem in general, where the input-output signals can be infinitely wide. In practice, solving this directly when the signals are very wide is typically computationally intractable.

Instead, we propose an approximate solution to \eqref{eq:mse_general} by considering small windows over the signals. To evaluate this analytically, we model \(X,~Y\) as stationary random signals and assume the model \(\bm\Phi\) is a CNN. Note that if $X$ is a random signal, then the model's output \(\bm\Phi(X)\) is another random signal. In Section \ref{sec:window} we provide a bound on the MSE in \eqref{eq:mse_general} when \(\bm\Phi\) is a CNN trained on small windows of the input-output signals. To facilitate this analysis we first define stochastic processes, stationarity, and CNNs.

\subsection{Stationary Stochastic Processes}
A stochastic process \cite{Gray09-ProbabilityRandom, Kallenberg21-Foundations, Shalizi07-AlmostNoneTheory} is a family of random variables \(\{X(t): \Omega \to S\}_{t \in \calI}\) parametrized by an index set \(\calI\). Each element in this family is a map from a sample space \(\Omega\) onto a measurable space \(S\). Equivalently, we can describe the process as a map \(X(t,\omega): \calI \times \Omega \to S\) from the cartesian product of the index set and the sample space onto the measurable space. In this context, a random signal is the corresponding random function, \(X := X(\cdot, \omega)\). For our purposes, these three representations are equivalent and we use them interchangeably.

We limit our discussion to real continuous stationary stochastic processes. In particular, consider two stochastic processes \(X: \reals \times \Omega \to \reals\) and \(Y: \reals \times \Omega \to \reals\). We assume that \(X\) and \(Y\) are jointly stationary, following definition \ref{def:jointly_stationary}. Simply put, two continuous random signals \(X\) and \(Y\) are jointly stationary if and only if all their finite joint distributions are shift-invariant.
\begin{definition}\label{def:jointly_stationary}
    Consider two real continuous stochastic processes \(\{X(t)\}_{t\in\reals}\) and \(\{Y(t)\}_{t\in\reals}\). The two processes are jointly stationary if and only if they satisfy \eqref{eq:jointly_stationary} for any shift \(\tau \in \reals\), non-negative \(n, m \in \mathbb{N}_0\), indices \(t_i, s_i \in \reals\), and Borel sets of the real line \(A_i, B_i \in \calB\).
    \begin{equation}
        \begin{split}\label{eq:jointly_stationary}
            &P(X(t_1) \in A_1, ..., X(t_n) \in A_n, ...,  Y(s_m) \in B_m)\\
            &=P(..., X(t_n+\tau) \in A_n, ..., Y(s_m+\tau) \in B_m)
        \end{split}
    \end{equation}
\end{definition}

Following \eqref{eq:mse_general}, we interpret \(X\) as a quantity that we can observe, while \(Y\) as something that we want to estimate. Numerous signals of interest such as financial \cite{Rinn15-Dynamics}, weather \cite{Michelangeli95-Weather} and multi-agent systems \cite{Owerko23-MultiTargetTracking} data can be represented by stationary or quasi-stationary processes. Therefore, it is clear that characterizing the performance of ML models for stationary signals is of paramount interest. In particular, we consider CNNs which themselves exhibit the same translational symmetries as stationary signals. Such analysis will allow us to further demystify their performance and provide guidelines on how to efficiently train CNNs.

\subsection{Convolutional Neural Networks}

CNNs are powerful architectures and oftentimes the tool of choice to approach the problem in \eqref{eq:mse_general}. The convolution operation is the cornerstone of the CNN architecture. Convolutions are shift-equivariant, which allows them to exploit stationarity in the signals. In its simplest form, a CNN is a cascade of layers, where the output of the \( l^\text{th} \) layer can be described by \eqref{eq:cnn}.
\begin{equation}\label{eq:cnn}
    x_l(t) = \sigma \left( \int h_l(s) x_{l-1}(t - s) ds \right)
\end{equation}
The output \(x_l\) at the \(l^\text{th}\) layer is obtained by convolving the input $x_{l-1}$ by a filter \(h_{l-1}\) and applying a pointwise nonlinearity \(\sigma_l\). Each filter is a continuous real function and we assume that it has finite support.

Let \(\bm\Phi(X; \calH)(t) := x_L(t) \) be the output of a CNN with \(L\) layers, input \(x_0(t) := X(t)\), and a set of filters \(\calH = \{ h_1,...,h_L\}\) as described by \eqref{eq:cnn}. Thus, the problem in \eqref{eq:mse_general} can be reformulated as finding a set of filters \(\calH\) that minimizes \(\Linf(\calH)\), the expected average squared error.
\begin{equation}\label{eq:mse_inf}
    \Linf(\calH) = \E{ \avg{ |\bm\Phi(X; \calH)(t) - Y(t)|^2 } }
\end{equation}
Evaluating \eqref{eq:mse_inf} is challenging since it requires data processing over all real numbers. Although real-world signals typically have finite support, in many applications the signals are too wide to realistically collect data or evaluate \eqref{eq:mse_inf}. For example, in Section \ref{sec:experiments}, we consider an optimization problem involving a large multi-agent system.

\section{Training on a window}\label{sec:window}

To overcome the aforementioned challenges we propose a transfer learning approach, where we train over small windows and execute the trained network in larger settings. Instead of minimizing \eqref{eq:mse_inf} directly, we can solve a simpler problem where we window the input-output signals \(X,~Y\). In this section, we summarize our theoretical result. It provides an upper bound for \(\Linf\) as described by \eqref{eq:mse_inf} in terms of the cost \(\Lfin(\calH)\) associated with \eqref{eq:mse_window}, which is easier to evaluate.

To characterize the behavior of CNNs in this context, we consider windows over the input and output signals of width \(\inW\) and \(\outW\), respectively; we restrict our attention to the case when \(\inW \ge \outW\). Denote a square pulse of width \(s \in \reals_+\) by \(\sqcap_s\), so that \(\sqcap_s(t) = 1\) whenever \(t \in [-s/2, s/2]\). \eqref{eq:mse_window} defines a new cost function.
\begin{equation}\label{eq:mse_window}
    \Lfin(\calH) = \frac{1}{\outW}
    \E{ \int_{-\outW/2}^{\outW/2} | \bm\Phi(\sqcap_\inW X; \calH)(t) - Y(t) |^2  dt }
\end{equation}
Unlike \eqref{eq:mse_inf}, this can be readily computed since the signals involved are time-limited, especially when \(\inW, \outW\) are small. Therefore optimizing the set of parameters \(\calH\) with respect to \(\Lfin\) is possible in practice. Our theoretical result quantifies the performance degradation when training on a small window, but evaluating on the whole signal. In particular, let \(\hat\calH\) be a candidate obtained by minimizing \eqref{eq:mse_window}. Given the following assumptions, we show that \(\hat\calH\) is an almost equally good solution \eqref{eq:mse_inf}.

\begin{assumption}\label{assume:jointly_stationary}
    The random signals \(X\) and \(Y\) are jointly stationary following definition \ref{def:jointly_stationary}.
\end{assumption}
\noindent This is the main assumption of our analysis. As mentioned earlier we analyze the performance of CNNs, since various data in high-impact domains are exactly or approximately stationary, and also such an analysis will also allow us to demystify the `black-box' architecture of CNNs. The remaining assumptions are very-mild and also typical for model analysis.

\begin{assumption}\label{assume:processes_bounded}
    The stochastic processes \(X\) and \(Y\) are bounded so that \( |X(t)| < \infty \) and \( |Y(t)| < \infty \) for all \(t\).
\end{assumption}
\noindent Boundedness of the signals is a sufficient condition for the existence and finiteness of some limits needed for the proof. However, the magnitude of the bound does not matter for the final result.

\begin{assumption}\label{assume:finite_filters}
    The filters \( h_l \in \calH \) are continuous with a finite width \(K\). That is \(h_l(t) = 0\) for all \(t \notin [K/2, K/2]\).
\end{assumption}
\begin{assumption}\label{assume:filters_bounded}
    The filters have finite L1 norms \(|| h_l ||_1 \le \infty\) for all \( h_l \in \calH \).
\end{assumption}
\noindent Assumptions \ref{assume:finite_filters} and \ref{assume:filters_bounded} are going to be satisfied in a typical implementation of a CNN which represents the filters as vectors -- or tensors when the signals are multi-dimensional.

\begin{assumption}\label{assume:lipschitz}
    The nonlinearities \( \sigma_l(\cdot) \) are normalized Lipshitz continuous (the Lipshitz constant is equal to 1).
\end{assumption}
\noindent Note that the majority of pointwise nonlinear functions used in deep learning, e.g., ReLU, Leaky ReLU, hyperbolic tangent, are normalized Lipshitz for numerical stability.

Theorem \ref{thm:stationary_bound} quantifies the difference between the mean squared error of a CNN trained according to \eqref{eq:mse_inf}, and executed on larger settings, according to \eqref{eq:mse_window}.
\begin{theorem}\label{thm:stationary_bound}
    Let \(\hat\calH\) be a set of filters which achieves a cost of \(\Lfin(\hat\calH)\) on the windowed problem as defined by \eqref{eq:mse_window} with an input window of width \(\inW\) and an output window width \( \outW \). Then the associated cost \( \Linf(\hat\calH) \) on the original problem as defined by \eqref{eq:mse_inf} is bounded by the following.
    \begin{equation}\label{eq:stationary_bound}
        \Linf(\hat\calH) \le \Lfin(\hat\calH) + H \frac{\outW + LK - \inW}{\outW} \text{var}(X)
    \end{equation}
    In \eqref{eq:stationary_bound}, \(H = \prod_l^L || h_l ||_1\) is the product of all the L1 norms of the CNNs filters, \(L\) is the number of layers, and \(K\) is the width of the filters.
\end{theorem}
The proof of Theorem \ref{thm:stationary_bound} is ommited due to space limitations. Theorem \ref{thm:stationary_bound} leverages the shift-equivariance property of CNNs to justify that CNNs are the suitable architecture for transfer learning. Since shift-equivariance is equivalent to joint stationarity of random signals, CNNs are able to exploit hidden regularities and translational symmetries in signals. In a nutshell, Theorem \ref{thm:stationary_bound} shows that the difference between the squared error of a CNN trained on a small setting and executed on a larger setting is bounded by a quantity that is affected by the variance of the input signal, number of layers, filter widths, and the size of the input and output windows. When $\inW=\outW+LK$, \eqref{eq:stationary_bound} reduces to \(\Linf(\hat\calH) \le \Lfin(\hat\calH)\). This special case occurs because for $\inW=\outW+LK$ there is no zero-padding at the intermediate CNN layers, eliminating border effects.

\section{Representing spatial problems as images}\label{sec:representing}
In this section, we explain how we represent spatial problems as images. Consider a machine learning task where the input is a set of positions \(\calX = \{ \bfx_n \mid \bfx \in \reals^2 \} \) and the output is another set of positions \( \calY = \{ \bfy_n \mid \bfx \in \reals^2 \} \). Our goal here is to represent these sets by images \( \bfX, \bfY \in \reals^{N \times N} \). After doing so, we can learn a mapping from \(\bfX\) to \(\bfY\) using a CNN and take advantage of Theorem \ref{thm:stationary_bound}.

\begin{figure*}[ht]
    \centering
    \includegraphics[width=\linewidth]{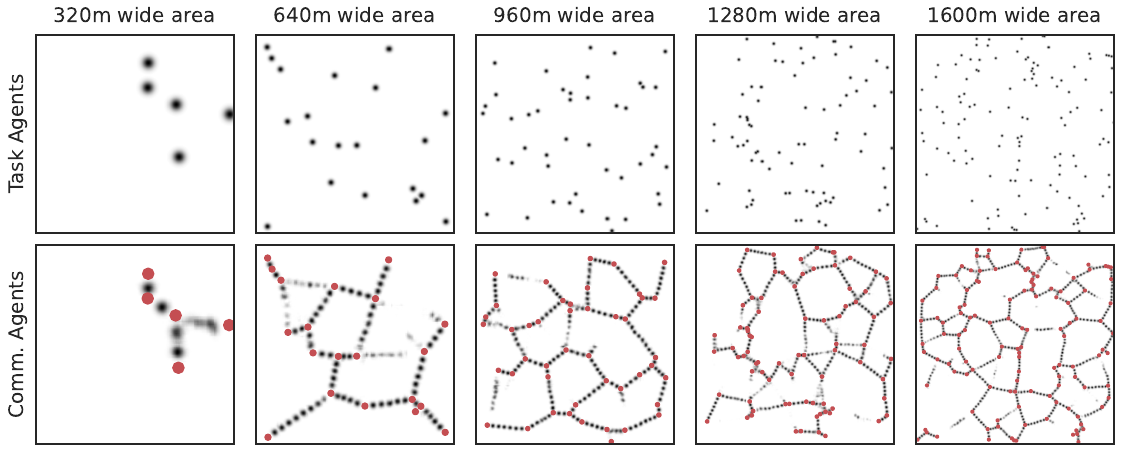}
    \caption{Example inputs and outputs to the CNN for the MID task at different window widths \( \inW = 320, 640, 960, 1280, 1600 \) but with a constant spatial resolution of \( \rho = 1.25 \) meters per pixel. The top row images represent the positions of the task agents. The bottom row images represent the estimated optimal positions of the communication agents by the CNN. Additionally, the task agent positions are marked in red on the bottom images.}
    \label{fig:mid_transfer}
\end{figure*}

We represent \( \calX \) by a superposition of Gaussian pulses with variance \(\sigma_x^2\). There is one pulse for each position \( \bfx_i \in \calX \). \eqref{eq:intensity_target} defines a real function that maps a position \(\bfx \in \reals^2\) to a real value, given the set of positions \( \calX \).
\begin{equation}\label{eq:intensity_target}
    X(\bfx; \calX ) := \sum_{i = 1}^{|\calX|} (2 \pi \sigma_x^2)^{-1}
    \exp(-\frac{1}{2\sigma_x^2}||\bfx-\bfx_{i}||^2_2)
\end{equation}
Next, we apply a square window of width \(\inW\) to \( X(\cdot; \calX) \) and sample with a spatial resolution \( \rho \) to obtain the matrix \( \bfX \).
\begin{equation}\label{eq:sample}
    \bfX_{ij} = X( \bmat{\rho i, \rho j}; \calX)
\end{equation}
\eqref{eq:sample} describes how the \(i,j^\text{th}\) pixel of \(\bfX\) is sampled for for \(i,j \in \{1,...,N\}\). The image width is rounded down to \( N = \lfloor \rho \inW \rfloor \). The images in the top row of Figure \ref{fig:mid_transfer} are example input images \( \bfX \) at different scales. Similarly, we can construct an image \( \bfY \) with a width of \(\outW\) meters from the set \( \calY \).

\section{Experimental results}\label{sec:experiments}
In this section, we test the performance of the proposed framework for mobile infrastructure on demand (MID). The primary goal in MID is to find the positions of a team of \emph{communication agents} that maximize wireless connectivity between a team of \emph{task agents}. The work in \cite{Mox20-MobileWirelessNetwork} formulated a convex optimization problem to find the optimal positions of the communication agents. However, this approach becomes computationally intractable as the number of agents grows. Instead, \cite{Mox22-Learning} proposed a data-driven approach, where a CNN is trained to imitate a convex optimization solution.

The authors of \cite{Mox22-Learning} represented the task and communication agent positions as images, following the procedure similar to Section \ref{sec:representing}. Hence, they were able to train a CNN with a fully convolutional encoder-decoder architecture on examples for two to six task agents uniformly distributed within a 320m window. The final model provided close to optimal configurations of communication agents. However, larger window sizes of task teams were never considered, which is what we focus on in this section.

Motivated by Theorem \ref{thm:stationary_bound}, we utilize a CNN to perform MID in large-scale settings. Specifically, we present our experimental results on zero-shot performance of the model proposed in \cite{Mox22-Learning} on MID tasks with up to 125 task agents and 562 communication agents, covering windows of up to 1600m.

We consider varying window widths of \(\inW = \) 320, 640, 960, 1280, and 1600 meters with \( |\calX| = \) 5, 20, 45, 80, and 125 task agents respectively. The number of task agents is proportional to the window area. We use the trained model provided by \cite{Mox22-Learning} from the associated GitHub repository, which was trained using a window size of \( \inW = \)320m. We perform no additional training or fine-tuning since we are only interested in evaluating the zero-shot performance.

At each window width \(\inW\) we evaluate the performance for 100 different random task agent configurations. The positions of the task agents are sampled uniformly within the square window to obtain a set \(\calX\). Using the procedure outlined in Section \ref{sec:representing} we represent the set of positions as an image \( \bfX \in \reals^{N \times N} \) where \(N = \lfloor \rho \inW \rfloor \). We use a spatial resolution of \( \rho = 1.25 \) meters per pixel and Gaussian kernel standard deviation \( \sigma_x = 6.4 \). The corresponding output of the CNN \( \bm\Phi(\bfX; \calH) \in \reals^{N \times N} \) is another image of the same size that represents the estimated optimal positions of the communication agents. Examples of the resulting CNN output images are shown in Figure \ref{fig:mid_transfer}. Qualitatively, the proposed transfer learning approach shows excellent performance in positioning the communication agents.

To quantify the performance of the proposed approach, we extract the set of the CNN estimated positions of the communication agents \( \hat\calY \) using Lloyd's algorithm as described by \cite{Mox22-Learning}. Then, we use Equation \eqref{eq:minimum_power} to calculate the power required to maintain a minimum communication rate between any two agents, assuming a path loss channel model \cite{Fink11-CommunicationTeams}. If two agents are separated by \(d\) meters, \eqref{eq:minimum_power} relates the power \(P(d)\) in mili-watts required for them to directly communicate at an expected communication rate \(R\). The rate is normalized to be a fraction of the total bitrate across the channel.
\begin{equation}\label{eq:minimum_power}
    P(d) = [\text{erf}^{-1}(R)]^2 \frac{P_{N_0} d^n}{K}
\end{equation}
In \eqref{eq:minimum_power} \(P_{N_0}\) is the noise power level in mili-watts, \(K\) is a constant based on the physics of the receivers, and \(n\) is known as the path loss exponent. In our simulations we use \(R = 0.5\), \(P_{N_0} = 1 \times 10^{-7}\), \(K = 5 \times 10^{-6}\) and \(n = 2.52\). Notice that \(P(d)\) is defined for every pair of agents and therefore defines a fully connected weighted graph between all agents. Consider the minimum spanning tree of this graph. The average edge weight of the tree is the per-edge power needed for communication.

\begin{table}
    \centering
    \caption{Average power per edge needed to maintain communication, the number of task agents, and the average number of communication agents for varying window width.}
    \label{tab:mid_power}
    \begin{tabular}{@{}rrrcc@{}}
        \toprule
        \multirow{2}{*}{Width (m)} & \multirow{2}{*}{Task Agents} & \multirow{2}{*}{Comm. Agents} & \multicolumn{2}{c}{Power (mW)}        \\
        \cmidrule{4-5}
                                   &                              &                               & Mean                           & STD  \\
        \midrule
        320                        & 5                            & 12.57                         & 16.60                          & 3.21 \\
        640                        & 20                           & 68.75                         & 18.30                          & 4.54 \\
        960                        & 45                           & 165.79                        & 18.27                          & 2.71 \\
        1280                       & 80                           & 308.85                        & 18.48                          & 2.06 \\
        1600                       & 125                          & 494.24                        & 18.03                          & 1.37 \\
        \bottomrule
    \end{tabular}
\end{table}

\begin{figure}
    \centering
    \includegraphics[width=0.85\linewidth]{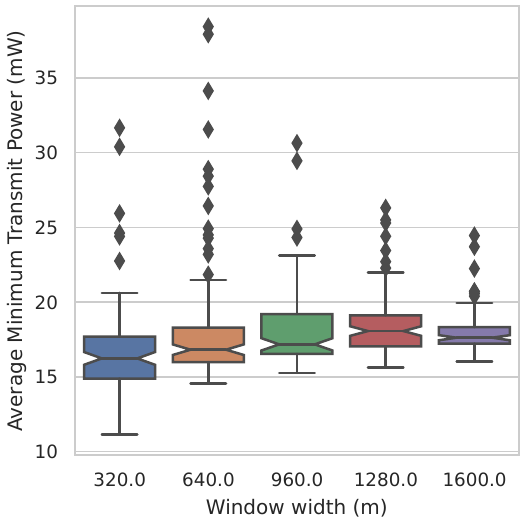}
    \caption{The distributions of the minimum transmitter power needed to maintain a normalized communication rate of at least 50\% between any two agents. The distributions for each window width are visualized by box plots, with notches representing a 95\% confidence interval for the estimate of the median.}
    \label{fig:mid_power}
\end{figure}

At each window width \(\inW\) we compute the power needed for each sampled of task agent configuration and the corresponding CNN prediction. As Table \ref{tab:mid_power} shows, there is a modest increase of 10.24\% in the power needed from \(\inW = 320\) to \(\inW = 640\) meters, but the power needed does not increase further after that. In fact, there is only an 8.61\% increase in power from \(\inW = 320\) to \(\inW = 1280\) meters. This suggests that further increases in the scale of the task, will not increase the needed power further.

Looking at Figure \ref{fig:mid_power}, the variance in the average power per edge decreases with the width. Partially, this is a consequence of the Central Limit Theorem, because in Figure \ref{fig:mid_power} we plot the distribution of an average over the edges in the minimum spanning tree. However, this is only part of the story, because the variance decreases faster than the square root of the number of edges. We think this is because border effects due to padding in the CNN have less impact at higher scales.

To summarize, the CNN gives us a good solution to the MID problem with linear time complexity respective to the area. In contrast, the convex optimization approach has polynomial complexity \cite{Mox22-Learning} as the number of agents is proportional to the area, which makes it computationally intractable for a large number of agents. Finally, the high performance of the CNN on the task is aligned with our derived theoretical bound.

\section{Conclusions}

This paper explores the use of CNNs to solve big spatial problems, which are too large to solve directly and to obtain training data. We present a novel theoretical result that expands our understanding of the link between shift-equivariance and CNN performance. Motivated by these results, we propose an approach to efficiently train a CNN to solve such tasks.  Specifically, a CNN can be trained on a small window of the signal and deployed on arbitrarily large windows, without a large loss in performance. We provide conditions that theoretically guarantee no loss of performance when using this method. To demonstrate this approach experimentally, we recast the MID problem as an image-to-image prediction task. In fact, the proposed approach is able to solve the MID problem for hundreds of agents, which was previously computationally intractable.

\bibliography{IEEEabrv,bib/settings,bib/journal}

\end{document}